\documentclass[conference]{IEEEtran}
\IEEEoverridecommandlockouts

\usepackage{amsmath, amssymb,amsfonts}
\usepackage{graphicx,subcaption}
\usepackage{cite}
\usepackage{url}
\usepackage{booktabs}
\usepackage{hyperref}
\usepackage{algorithm}
\usepackage{algorithmic}
\usepackage{float}
\usepackage[utf8]{inputenc}
\usepackage{textgreek}
\usepackage{multirow}

\begin{document}

\title{Comparative Analysis of Ant Colony Optimization and Google OR-Tools for Solving the Open Capacitated Vehicle Routing Problem in Logistics}

\author{
    \IEEEauthorblockN{Assem Omar}
    \IEEEauthorblockA{\textit{Faculty of Computer Science}\\
    MSA University\\
    Giza, Egypt\\
    asem.omar2@msa.edu.eg}
    \and
    \IEEEauthorblockN{Youssef Omar}
    \IEEEauthorblockA{\textit{Faculty of Computer Science}\\
    MSA University\\
    Giza, Egypt\\
    youssef.mohamed81@msa.edu.eg}
    \and
    \IEEEauthorblockN{Marwa Solayman}
    \IEEEauthorblockA{\textit{Faculty of Computer Science}\\
    MSA University\\
    Giza, Egypt\\
    mmsolayman@msa.edu.eg}
    \and
    \IEEEauthorblockN{Hesham Mansour}
    \IEEEauthorblockA{\textit{Faculty of Computer Science}\\
    MSA University\\
    Giza, Egypt\\
    hhmansour@msa.edu.eg}
}

\IEEEoverridecommandlockouts
\IEEEpubid{\makebox[\columnwidth]{ 979-8-3315-0185-3/25/\$31.00 ©2025 IEEE \hfill}
\hspace{\columnsep}\makebox[\columnwidth]{ }}
\maketitle
\IEEEpubidadjcol

\begin{abstract}
In modern logistics management systems, route planning requires high efficiency. The Open Capacitated Vehicle Routing Problem (OCVRP) deals with finding optimal delivery routes for a fleet of vehicles serving geographically distributed customers, without requiring the vehicles to return to the depot after deliveries. The present study is comparative in nature and speaks of two algorithms for OCVRP solution: Ant Colony Optimization (ACO), a nature-inspired metaheuristic; and Google OR-Tools, an industry-standard toolkit for optimization. Both implementations were developed in Python and using a custom dataset. Performance appraisal was based on routing efficiency, computation time, and scalability. The results show that ACO allows flexibility in routing parameters while OR-Tools runs much faster with more consistency and requires less input. This could help choose among routing strategies for scalable real-time logistics systems.
\end{abstract}


\section{Introduction}

Beyond typical road-based logistics, efficiently routing vehicles with capacity limits through open fields presents a unique challenge\cite{praveen2022vehicle}. With the advent of e-commerce and last-mile delivery services, the Open Capacitated Vehicle Routing Problem (OCVRP) has been garnering serious attention in the literature \cite{10486417}. Unlike the traditional CVRPs, the OCVRP does not make the vehicles return to the depot after completing deliveries; hence, it more accurately models scenarios wherein rented vehicles with one-way operations are considered \cite{ibrahim2020capacitated}.

OCVRPs are formed where a fleet provides service to many customers that have specific demands with routes that minimize total distance traveled, always upholding capacity restraints on the vehicle\cite{gasset2024route}. This variant presents a realistic context concerning modern logistics systems, especially in urban settings in which the delivery personnel could perhaps switch transport means after fulfilling the route \cite{10797158}. As e-commerce has expanded, the question of efficient routing solutions has only increased since they serve the dual purpose of maintaining competitive advantage while keeping operational costs in check\cite{mehmood2021does}.

Tackling complex scenarios with lots of rules and restrictions can overwhelm traditional optimization methods when applied to real-world scenarios \cite{muriyatmoko2024heuristics}. As the delivery networks grow beyond 50-100 points, exact methods become computationally prohibitive and would necessitate the use of efficient heuristic methods \cite{10221809}. In this study, we examine two algorithmic paradigms for their use in solving OCVRP:

\textbf{Ant Colony Optimization (ACO)}- a nature-inspired metaheuristic that would model the foraging behavior of an ant colony through pheromone-based communication. It had showed good performance in routing problems and exploring-exploiting in-complex search space \cite{10091367, 9757954}. Recently, improvements have targeted stagnation occurring at an initial stage and promoted performance in multi-objective settings \cite{doi:10.3233/KES-160335}. 

\textbf{Google OR-Tools} - an industrial-grade optimization toolkit covering constraint programming, metaheuristics, and mathematical optimization as parts of the improvised package. OR tools empower exact methods within guided local search and various construction heuristics for considerably robust performance at large-scale problems while their implementation requirements are simplified \cite{muriyatmoko2024heuristics, 10797158}.

The parameters for our comparison are not only based on quality of solutions, but also on complexity of implementations, efficiency of computation, parameter tuning requirements, and solution stability. Especially so for small to medium-sized logistics companies that have very limited algorithm development resources \cite{ibrahim2020capacitated}. Implementing both approaches in Python on the same real-world datasets helps provide insights on the performance of logistic software developers interested in getting efficient routing solutions in resource-constrained environments \cite{10221809}.

\section{Problem Description: OCVRP}

The OCVRP has the following characteristics:
\begin{itemize}
    \item All vehicles start from a single depot.
    \item A certain number of customers have fixed demands exactly 1.0 demand per location.
    \item Each vehicle belongs to a fleet with a limited capacity .
    \item Each vehicle must visit customers without exceeding its capacity.
    \item \textbf{Not returning} to the depot after delivery.
    \item The goal is to minimize the total distances traveled by all vehicles.
    \item With 100\% capacity utilization, meaning that all vehicles are used to their maximum capacity 
\end{itemize}

This distinction, allows the vehicle to finish at its last customer, means that there can be quite a different optimization scenario from the traditional CVRP and that it presents a more realistic context which would be available from a one-way delivery service, particularly relevant in modern last-mile delivery systems \cite{10486417}.

\section{Related Work}

Studies on Vehicle Routing Problems
Somewhat influenced by logistics and transportation optimization, the Capacitated Vehicle Routing Problem has reputedly received substantial attention within the research community. For instance, Praveen et al. \cite{praveen2022vehicle} presented a very exhaustive review of CVRP, wherein the authors discussed several approaches for its solution, and applications were examined with respect to practicality in a logistics environment. Within their study, they proposed a classification of the problem formulations based on varying constraints and objectives and emphasized the need to find a balance between solution quality and computational efficiency.

The Open Vehicle Routing Problem was the main concentration for Gasset et al. \cite{gasset2024route}, who proposed formulations and solution techniques that did not require the vehicle to return to the depot. They argued that the OVRP better accounts for actual logistics situations because unique rentals could involve one-way trips, or people could change their means of transport after completing their delivery routes.

Ibrahim et al. \cite{ibrahim2020capacitated} applied some column generation techniques in CVRP-inspired reinforcement learning, thus confirming that hybrid methods can drive in an outcome better than methods by themselves. This method appears particularly encouraging for small to medium logistics operators' interventions, considering they work often with limited computational resources.

B. Ant Colony Optimization in Routing Applications
Ant Colony Optimizations have been proved applicable to a variety of routing problems. Liu et al. \cite{9363800}, by applying ACO to the logistics distribution route optimization, validated its application in multi-objective scenarios where distance and time constraints need consideration. They stressed ACOs exploration and exploitation capability in the complex search space.

Temperature-controlled delivery presents various challenges to cold chain logistics distribution as described in Wang \cite{10075989}, who proposed an ameliorated algorithm based on the ACO method with these challenges in mind. The modifications of the standard ACO add adaptive parameter setting and local search strategies to some extent contribute to solution quality improvements.

In Xuyou et al. \cite{10797158}, the ACO method was advanced and augmented for logistics and distribution path optimization techniques by embedding time variables in the heuristic information. The pheromone evaporation rates were adjusted dynamically with different ratings, helping to postpone the premature convergence towards a suboptimal solution.

Aggarwal et al. \cite{10486417} proposed ant-like algorithms for the optimization of last-mile delivery and illustrated that ACO was particularly suited for dense urban delivery networks. The tests indicated that the algorithm worked well for scenarios of 50 to 100 delivery points, which was in line with the scale of this study.

C. Algorithmic Improvements and Multi-Objective Approaches
Enhancement of ACO algorithms to tackle the problems of premature convergence and stagnation has been the research focus for some time now. Deng et al. \cite{10091367} designed a multi-objective ant colony algorithm for path planning problems with time windows, with one objective being weighted against others using Analytic Hierarchy Process. Their algorithm performed best under strict time constraints.

The model of optimal travel route optimization that Zhang and Sun \cite{10221809} proposed is based on ACO and includes a stability analysis. The approach has particularly focused on the scalability issues of problems by making instances grow beyond a range of 50-100 points.

Niu \cite{9788179} applied topology optimization with ACO algorithms in tourism route planning and produced an entirely multi-objective scheme for considering cost and user satisfaction as criteria. The topology-based modifications enabled a better search process in complicated geographical spaces.

Huang et al. \cite{9757954} addressed reliability in multi-agent path planning using an augmented ACO algorithm with forecasting mechanisms to avoid agent-agency conflicts. The proposed approach is more suitable for the fleet management scenario, which involves synchronizing multiple vehicles at once.

Comparative studies of the various optimization methods are invaluable to practitioners. Muriyatmoko et al. \cite{muriyatmoko2024heuristics} compared different heuristics and metaheuristics of CVRP solutions, including ACO and operations research-based methods. Their finding suggested that although metaheuristic ACO is flexible, industrial-grade optimization tools, like OR-Tools, can give steadier results with much less tuning and parameter consideration.

Indravattana and Silawan \cite{10147562} studied load balancing among ACO multi-depot pickup and delivery problems and then compared it with conventional methods. The study highlighted the significance of workload balancing among vehicles, which is one of the primary considerations in normal logistics operations.

According to Jiang and Sun \cite{10873146}, ACO was used for police patrol line planning and compared it with some other optimization techniques. The study showed that ACO can model quite complex constraints in patrol routing with computational efficiency.

Sari et al. \cite{10462744} reviewed all path planning methods based on the ant colony algorithm and compared their various types with their applications. Their review also stated that ACO is applicable in many routing contexts with different constraints and objectives.

The literature shows that while ACO provides adaptability and flexibility to handling complex constraints, it is too inefficient to implement industrial solvers within these bounds. Our study builds upon these attributes by comparing ACO and Google OR-Tools for solving OCVRP in logistics. Our focus is on practical implementation aspects and performance characteristics of the two approaches.

\section{Methodology}

\subsection{Experimental Setup}
For the purpose of algorithmic comparison, every experiment was conducted on a single machine (AMD Ryzen 6800H CPU @ 3.2~GHz, 16~GB DDR5 RAM). Everything was implemented in Python, utilizing NumPy for fast numerical compute and plotting via Matplotlib. Performance measures are taken in wall-clock time rather than CPU time to better represent a real-world application's performance, including time expenditure for other computational overheads and latencies of memory access, but excluding time spent for initial setups like loading data into RAM, constructing the distance matrix, and initializing the solver, additionally both algorithms used single threading to ensure a fair comparison between them.

\subsection{Dataset Collection and Processing}
\textbf{1. Data Source}
\begin{itemize}
  \item Coordinates of real cities located within Greater Cairo, Egypt (five cities).
  \item OpenStreetMap (OSM) gives actual road network data.
  \item OSM-Pre-computed distance matrix.
\end{itemize}

\textbf{2. Data Storage Formats}
\begin{itemize}
  \item JSON file with location coordinates (Latitude, Longitude), demands, and time windows.
  \item Vehicle definitions with capacity, fixed costs, and time windows.
  \item Distance matrix accounted in a NumPy \texttt{.npz}.
  \item Clustered Variants (\texttt{Greater\_cairo\_100\_10.json}) where locations are pre-grouped for improved routing.
\end{itemize}

\subsection{Ant Colony Optimization}

Ant Colony Optimization (ACO) algorithms mimic ant foraging behavior, whereby artificial ants build solutions for the OCVRP, probabilistically generating routes that start at the depot and end with the last customer within the proper capacity constraints. The algorithm uses pheromone trails to encode information concerning the quality of paths already traveled, while heuristic information biases it towards shorter routes.

The ACO algorithm for OCVRP can be defined as follows:

\textbf{1. Solution Construction}

Each ant of the system would construct a route by assigning customers to vehicles and building routes. At every step, the ant located at \(i\) chooses the next location \(j\) from the set of unvisited customers \(N_i\), under the constraints of capacity. The choice for a particular \(j\) is then determined by a mixture of greedy and random decisions, parameterized by \(q_0\); where \(q_0 \in [0, 1]\):
\begin{itemize}
  \item With probability \(q_0\), the ant chooses the next location greedily:
  \begin{equation}
    j = \arg\max_{l \in N_i} \bigl([\tau_{il}]^\alpha \cdot [\eta_{il}]^\beta\bigr),
  \end{equation}
  \item With probability \(1 - q_0\), the ant chooses the next location probabilistically:
  \begin{equation}
    P_{ij} = 
    \begin{cases}
      \dfrac{[\tau_{ij}]^\alpha \cdot [\eta_{ij}]^\beta}{\sum_{l \in N_i} [\tau_{il}]^\alpha \cdot [\eta_{il}]^\beta}, & j \in N_i,\\
      0, & \text{otherwise},
    \end{cases}
  \end{equation}
\end{itemize}
where \(\tau_{ij}\) is the pheromone level on edge \((i,j)\), \(\eta_{ij} = 1/d_{ij}\) is the heuristic information (\(d_{ij}\) is the distance), \(\alpha\) is the pheromone influence, and \(\beta\) is the heuristic influence.

\textbf{2. Pheromone Update}

After all ants construct solutions, pheromone trails are updated:

\begin{itemize}
  \item \textit{Evaporation}: 
  \begin{equation}
    \tau_{ij} \leftarrow (1 - \rho)\,\tau_{ij},
  \end{equation}
  \item \textit{Deposition}:
  \begin{figure}[H]
      \begin{equation}
      \centering
      \includegraphics[width=1\linewidth]{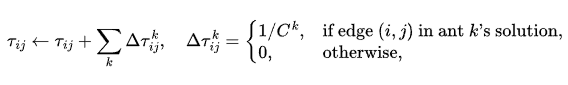}
      \label{fig:enter-label}
     \end{equation}

  \end{figure}
\end{itemize}
with \(\rho\) the evaporation rate and \(C^k\) the total distance of ant \(k\)'s solution.

\textbf{3. Local Search}

To enhance solution quality while meeting capacity constraints, we perform 2-opt local search on each route.

\textbf{4. Parameters}

\textit{Exploitation:} 
\begin{itemize}
  \item \(\alpha = 2.5\): This value indicates very high exploitation, as it puts pheromone trails on large influences so that the ants would rather follow previously discovered successful routes than explore new ones.
  \item \(\beta = 1.0\): Small influences of heuristic information (i.e., distance) allow ants to totally exceed greedy, distance-based decisions and rely more on learned information.
  \item \(\rho = 0.1\): A very low evaporation rate so that good pheromone trails last longer in an effort to maintain recollection of high-quality routes.
  \item \(q_0 = 0.9\): The high probability of making a greedy choice for the best edge intensifies exploitation of the known good path.
  \item iterations = 150, ants = 40. 
\end{itemize}

\textit{Exploration:}
\begin{itemize}
  \item \(\alpha = 0.2\): The influence of pheromone is weak and hence their tendency to follow established trails is reduced which promotes the exploration of a new path.
  \item \(\beta = 3.0\): An ant that gets heuristic information considers this over any pheromone level, hence distance-influenced decisions help ants in finding shorter edges.
  \item \(\rho = 0.7\): The higher the evaporation rate, the more quickly pheromones on all routes are decreased, thus preventing faster convergence into suboptimal solutions.
  \item \(q_0 = 0.1\): Low probability of being greedy ensures that more of the choices are made in a probabilistic way, which supports diverse route sampling.
  \item iterations = 150, number of ants = 40.
\end{itemize}

\textbf{5. Stagnation Handling}

In case of a stagnation that lasts for 20 iterations, its pheromone trails are reset to initial value with an exception of the best solution \cite{doi:10.3233/KES-160335}. Also, a safety mechanism is in place to prevent potential infinite loops in highly constrained cases through the implementation of a maximum attempt counter.

\subsection{Google OR-Tools Implementation}

Google's OR-Tools is a powerful and open-source platform suited to solving routing problems using constraint programming and guided local search. Here, advanced mathematical optimization techniques have been seamlessly woven with heuristics applied to vehicle routing problems. Our implementation was done in Python using ortools.constraint\_solver.pywrapcp.

\begin{itemize}
    \item \textbf{Routing Model Creation} 
    \begin{itemize}
        \item Where a mathematical creation of a graph representation is defined with nodes as locations, and the map between indices in the model and solver index is internally built. 
    \end{itemize}

    \item \textbf{Distance Callback Registration} 
    \begin{itemize}
        \item Register a distance calculation with numerical stabilization. 
        \item It sets the evaluator of costs for arc traverse between nodes. 
    \end{itemize}

    \item \textbf{Capacity Constraint Modeling} 
    \begin{itemize}
        \item Demand callback is registered at each location.
        \item Capacity dimension is created with vehicle-specific limits.
        \item Track total loads in the routes. 
    \end{itemize}

    \item \textbf{Open Route Configuration} 
    \begin{itemize}
        \item The depot is the starting point for the vehicles, but routes may end with any customer. 
        \item Instead of capturing this with an explicit constraint, such behavior is best implemented on the basis of extraction from solutions. 
    \end{itemize}

\item \textbf{Search Strategy Configuration} 
    \begin{itemize}
        \item Tries to apply a variety of construction heuristics sequentially~\cite{surana2019benchmarking}:
        \begin{itemize}
            \item \texttt{PATH\_CHEAPEST\_ARC}: A greedy sequential insertion strategy that builds routes by repeatedly adding the closest unvisited node to the end of the path.
            \item \texttt{PARALLEL\_CHEAPEST\_INSERTION}: This heuristic builds many routes at the same time: each customer is inserted in the position which offers the lowest increase in distance cost from all the routes. 
            \item \texttt{SAVINGS}: The Clarke \& Wright savings algorithm entails assigning each customer to its own route and iteratively combining the routes that yield the greatest distance savings. 
            \item \texttt{AUTOMATIC}: This allows OR-Tools to choose the most applicable construction heuristic based on the problem features and size of the instance. 
        \end{itemize}
        \item Guided Local Search is then applied as a metaheuristic to refine the solution, while escaping from local optima by dynamically changing the objective function using penalties for edges that are often used. 
    \end{itemize}

    \item \textbf{Solution Execution \& Extraction}     
    \begin{itemize}
        \item Solves the model for given parameters and a time limit. 

        \item Extracts routes based upon variable assignment. 

        \item Calculate the final metrics and delete empty routes. 

        \item The solutions are stored in JSON for visualization and analysis.
    \end{itemize}
\end{itemize}

\section{Evaluation and Results}
The algorithms were assessed on different datasets with different customer counts. The evaluation metrics include:

\begin{itemize}
    \item \textbf{Total Route Distance} (in Kilometers): How efficiently did each algorithm reduce the overall travel distance?
    
    \item \textbf{Computation Time} (in Seconds): How much time did the algorithms take to provide a best solution?
    
    \item \textbf{Constraint Satisfaction}: How well did each algorithm respect the vehicle capacity constraints?
\end{itemize}

Note: All vehicles met their capacity constraints, ensuring 100\% capacity utilization.     

\begin{figure}[H]
    \centering
    \includegraphics[width=1\linewidth]{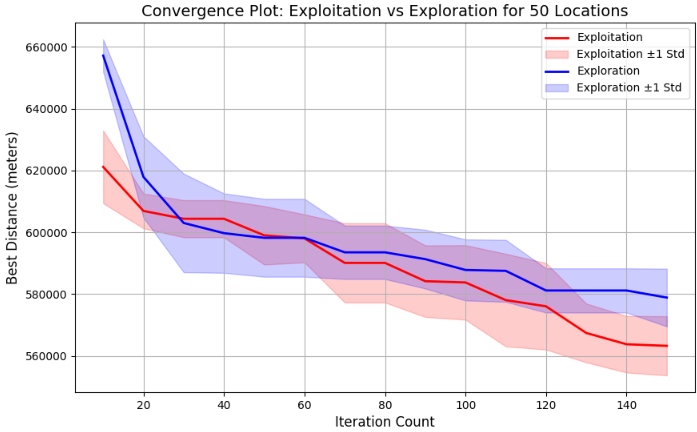}
    \caption{50-Location Convergence Plot}
    \label{fig:1}
\end{figure}

\textbf{Convergence Plot (50 locations)}:  
The \textbf{Exploitation} strategy showed a sudden drop in the first 50 iterations to approximately 580 km, followed by slow progress toward 560 km by 150 iterations. The standard deviation band remained relatively narrow, indicating low variability.

The \textbf{Exploration} strategy began at around 650 km and steadily decreased to 580 km over 150 iterations. The standard deviation band was slightly wider during iterations 30–40, reflecting higher variability as ants explored more aggressively, but it narrowed by iterations 130–140, suggesting stabilized performance in smaller problem instances (50 locations), as shown in Fig.~\ref{fig:1}.

\begin{figure}[H]
    \centering
    \includegraphics[width=1\linewidth]{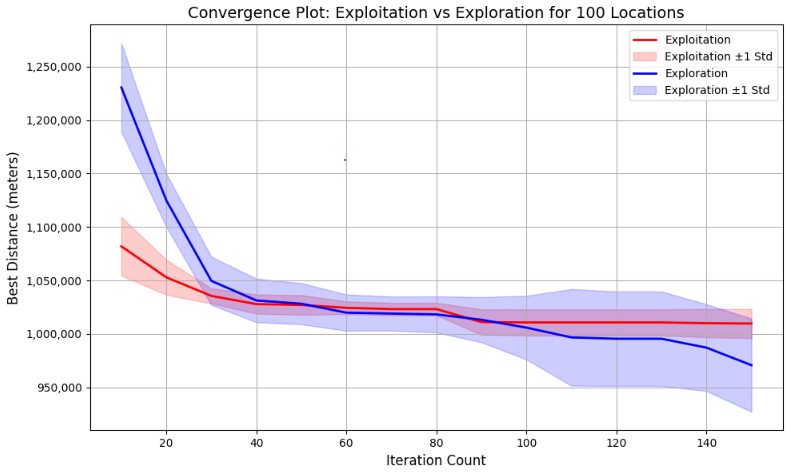}
    \caption{100-Location Convergence Plot}
    \label{fig:2}
\end{figure}

\textbf{Convergence Plot (100 locations)}:  
\textbf{Exploitation} starts at approximately 1,200 km, drops to approx.~1,050 km within 50 iterations, then flattens to approx.~1,000 km by iteration 150. The standard deviation band is narrow, showing that exploitation escaped local optima but was still less efficient in larger problem instances.  
\textbf{Exploration} starts at approx.~1,250 km and decreases steadily to approx.~950 km by iteration 150. The standard deviation band is wider, showing more variability in the final results, as shown in Fig.~\ref{fig:2}.

\begin{figure}[H]
    \centering
    \includegraphics[width=1\linewidth]{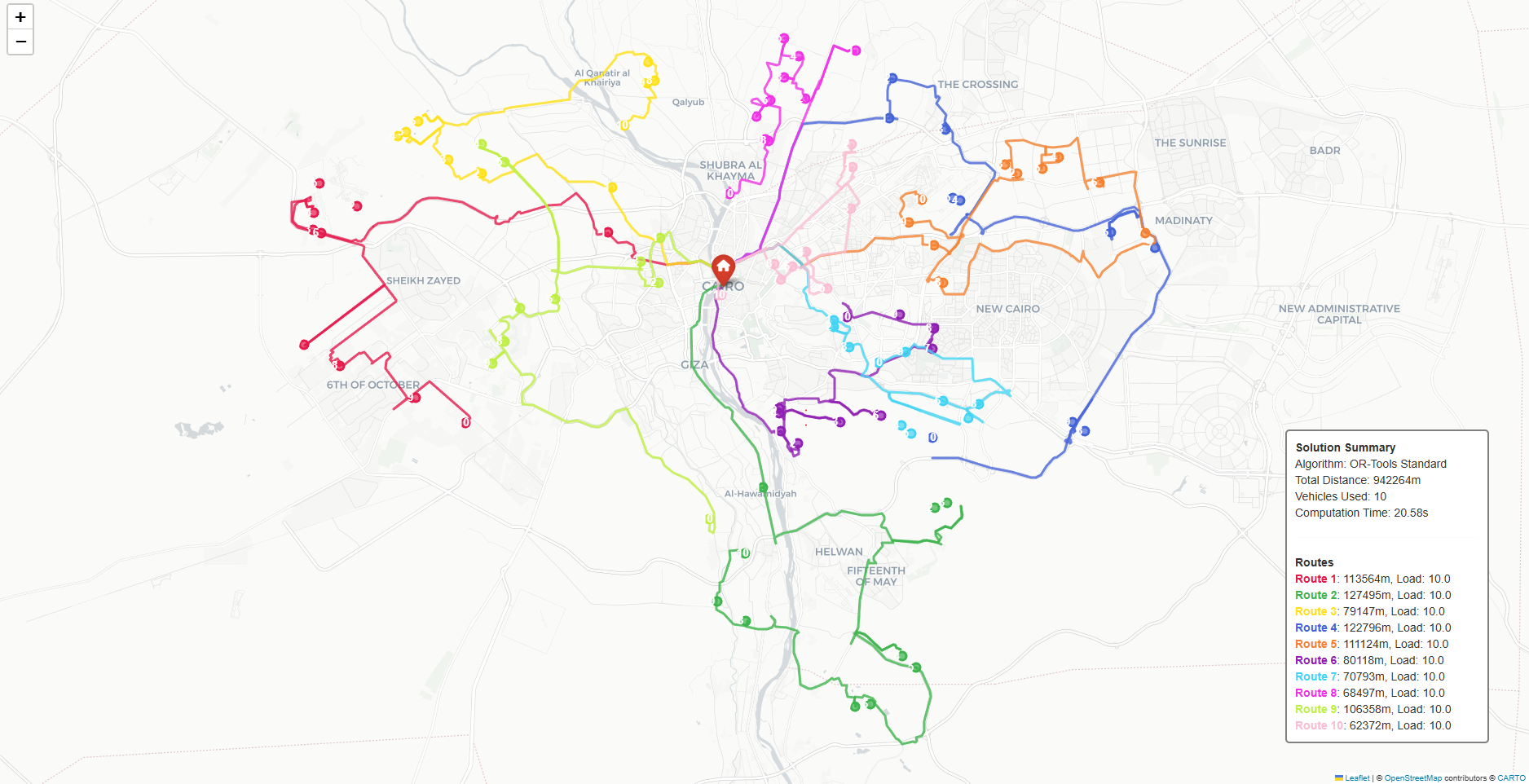}
    \caption{OR-Tools Map Visualization}
    \label{fig:3}
\end{figure}

\textbf{Route Visualization}:  
Fig.~\ref{fig:3} shows a visualized map of the final solution generated using OR-Tools. It illustrates how efficiently the vehicles were routed and how well the capacity and location constraints were handled in practice.

\begin{table}[htbp]
\caption{Performance Comparison: ACO vs. OR-Tools for OCVRP}
\centering
\small
\begin{tabular}{@{}l@{\;}l@{\;}r@{\;}r@{\;}r@{}}
\toprule
\textbf{Case} & \textbf{Metric} & \multicolumn{1}{c}{\textbf{ACO}} & \multicolumn{1}{c}{\textbf{ACO}} & \multicolumn{1}{c}{\textbf{OR-}} \\
\textbf{(n)} & & \multicolumn{1}{c}{\textbf{(Explor.)}} & \multicolumn{1}{c}{\textbf{(Exploit.)}} & \multicolumn{1}{c}{\textbf{Tools}} \\
\midrule
\multirow{2}{*}{50} & Dist. (km) & 578.9 $\pm$ 9.3 & 563.2 $\pm$ 9.6 & 590.0 \\
& Time (s) & 42.6 $\pm$ 9.0 & 101.1 $\pm$ 9.0 & 4.4 \\
\midrule
\multirow{2}{*}{100} & Dist. (km) & 970.8 $\pm$ 43.5 & 1009.8 $\pm$ 13.6 & 942.3 \\
& Time (s) & 308.1 $\pm$ 39.4 & 291.0 $\pm$ 25.6 & 9.6\\
\bottomrule
\end{tabular}
\label{tab:performance1}
\end{table}

The table reports the mean values along with their standard deviations ($\pm$~std), providing insights into both the average performance and the variability of the results. To ensure reproducibility, each approach (\textbf{Exploitation} and \textbf{Exploration}) was tested 10 times.  
Though the distance results from OR-Tools were deterministic (and hence no averaging was needed), the computational time still varied a bit (±0.13 seconds) in 100 locations and (±0.04 seconds) in 50 locations.

\vspace{1em}

Based on our study, we came to the following results:
 
\begin{itemize} 
  \item \textbf{Usability}: An easy-to-use API is available in OR-Tools that requires less development effort than writing full ACO code by hand. 
  \item \textbf{Configurability}: ACO allows fine tuning of parameters and routing behavior to perfectly fit specific requirements. 
  \item \textbf{Scalability}: OR-Tools is much more favorable than ACO with respect to scaling for very large datasets \cite{ibrahim2020capacitated}. 
  \item \textbf{Flexibility}: ACO works much better in dynamic or very constrained situations, guiding decisions through pheromone trails. 
\end{itemize}

\section{Conclusion and Future Work}

A comparative study on Ant Colony Optimization and Google OR-Tools for the solution of the Open Capacitated Vehicle Routing Problem brings out several value-added insights from a software engineering point of view. OR-Tools outperformed ACO both in computational speed and the quality of solutions and scalability across different sizes of problems while requiring very few parameter tuning. The implementation was especially efficient with a 20-30x speed-up for ACO computations while giving as good as or superior solution quality. Customization and transparency in solution development were brought by ACO's pheromone trail mechanism's provision for a more profound insight into the solution process.

In particular, OR-Tools turns out to be a suitable software to implement in levels of resource constraints, among small to medium logistics organizations, where algorithm development sources are limited \cite{ibrahim2020capacitated}. This is mainly because of the performance and implementation simplicity for OR-Tools. Such organizations wanting any kind of specialized routing or wishing to add specialized knowledge into the routing process would find the configurability of ACOs through parameter settings \cite{9757954} beneficial.

For example, for small to medium logistics organizations where the available resources for algorithm development are limited, such as in practical implementations under the constraint of resources, OR-Tools proves to be the best possible solution owing to its good performance and ease of implementation requirements. ACO would be advantageous to an organization which needs more specialized routing behavior and would like to add some specific domain knowledge into the routing process since it can be configured through parameter settings \cite{9757954}.

\noindent Future research directions may include:

Incorporation of dynamic rerouting capabilities with real-time traffic data so as to improve adaptability to urban delivery scenarios, as further developed from the ant-inspired approaches for last-mile delivery presented by Aggarwal et al. \cite{10486417}.

Generation of hybrid approaches that combine the exploratory capabilities of the ACO with the efficaciousness of OR-Tools in order to combat premature convergence, especially with regard to complex clustered routing scenarios as shown in the most recent studies in logistics distribution path optimization \cite{10797158}.

Creation of cloud-based OCVRP real-time solutions that utilize industrial-grade solvers for processing large dynamic routing problems and with regard to computational challenges encountered by Muriyatmoko et al. \cite{muriyatmoko2024heuristics}.

Extension towards multi-objective optimization frameworks such as route length, vehicle utilization, environmental impact, and time windows under the system of Deng et al. \cite{10091367} and tourism route optimization-related research \cite{9788179}.

Load balancing mechanisms among various vehicles as an aspect of equal distribution of deliveries with particular concern to dense urban environments would also benefit Indravattana and Silawan \cite{10147562}. 

Ultimately, these are the expected contributions to closing the gap between the theoretical development of algorithms and modern logistics systems' implementation realities, specifically regarding scalability issues for problem instances with delivery points in excess of 50-100 \cite{10221809}.

\bibliographystyle{IEEEtran}

\begin{thebibliography}{10}
\providecommand{\url}[1]{#1}
\csname url@samestyle\endcsname
\providecommand{\newblock}{\relax}
\providecommand{\bibinfo}[2]{#2}
\providecommand{\BIBentrySTDinterwordspacing}{\spaceskip=0pt\relax}
\providecommand{\BIBentryALTinterwordstretchfactor}{4}
\providecommand{\BIBentryALTinterwordspacing}{\spaceskip=\fontdimen2\font plus
\BIBentryALTinterwordstretchfactor\fontdimen3\font minus \fontdimen4\font\relax}
\providecommand{\BIBforeignlanguage}[2]{{%
\expandafter\ifx\csname l@#1\endcsname\relax
\typeout{** WARNING: IEEEtran.bst: No hyphenation pattern has been}%
\typeout{** loaded for the language `#1'. Using the pattern for}%
\typeout{** the default language instead.}%
\else
\language=\csname l@#1\endcsname
\fi
#2}}
\providecommand{\BIBdecl}{\relax}
\BIBdecl

\bibitem{praveen2022vehicle}
V.~Praveen, P.~Keerthika, G.~Sivapriya, A.~Sarankumar, and B.~Bhasker, ``Vehicle routing optimization problem: a study on capacitated vehicle routing problem,'' \emph{Materials Today: Proceedings}, vol.~64, pp. 670--674, 2022.

\bibitem{10486417}
A.~Aggarwal, P.~Ghosh, K.~Sharma, S.~Sharma, M.~Raj, and M.~M. Ali, ``Ant-inspired route optimization for last mile delivery,'' in \emph{2024 IEEE International Conference on Computing, Power and Communication Technologies (IC2PCT)}, vol.~5, 2024, pp. 366--371.

\bibitem{ibrahim2020capacitated}
A.~Ibrahim, J.~Ishaya, N.~Lo, and R.~Abdulaziz, ``Capacitated vehicle routing problem with column generation and reinforcement learning techniques,'' \emph{Open J. Discret. Appl. Math}, vol.~3, no.~1, pp. 41--54, 2020.

\bibitem{gasset2024route}
D.~Gasset, F.~Paillalef, S.~Payac{\'a}n, G.~Gatica, G.~Herrera-Vidal, R.~Linfati, and J.~R. Coronado-Hern{\'a}ndez, ``Route optimization for open vehicle routing problem (ovrp): A mathematical and solution approach,'' \emph{Applied Sciences}, vol.~14, no.~16, p. 6931, 2024.

\bibitem{10797158}
S.~Xuyou, Y.~Shuyan, and S.~Suisui, ``Research on optimization of logistics and distribution paths based on improved ant colony algorithm,'' in \emph{2024 IEEE 6th International Conference on Power, Intelligent Computing and Systems (ICPICS)}, 2024, pp. 1705--1710.

\bibitem{mehmood2021does}
T.~Mehmood, ``Does information technology competencies and fleet management practices lead to effective service delivery? empirical evidence from e-commerce industry,'' \emph{International Journal of Technology Innovation and Management (IJTIM)}, vol.~1, no.~2, pp. 14--41, 2021.

\bibitem{muriyatmoko2024heuristics}
D.~Muriyatmoko, A.~Djunaidy, and A.~Muklason, ``Heuristics and metaheuristics for solving capacitated vehicle routing problem: An algorithm comparison,'' \emph{Procedia Computer Science}, vol. 234, pp. 494--501, 2024.

\bibitem{10221809}
L.~Zhang and P.~Sun, ``An optimal travel route optimization model based on ant colony optimization algorithm,'' in \emph{2022 5th Asia Conference on Machine Learning and Computing (ACMLC)}, 2022, pp. 105--110.

\bibitem{10091367}
C.~Deng, J.~Lin, and L.~Chen, ``A multi-objective ant colony algorithm for the optimization of path planning problem with time window,'' in \emph{2022 18th International Conference on Computational Intelligence and Security (CIS)}, 2022, pp. 351--355.

\bibitem{9757954}
S.~Huang, D.~Yang, C.~Zhong, S.~Yan, and L.~Zhang, ``An improved ant colony optimization algorithm for multi-agent path planning,'' in \emph{2021 International Conference on Networking Systems of AI (INSAI)}, 2021, pp. 95--100.

\bibitem{doi:10.3233/KES-160335}
\BIBentryALTinterwordspacing
A.~Byerly and A.~Uskov, ``A novel approach to avoiding early stagnation in ant colony optimization algorithms,'' \emph{International Journal of Knowledge-Based and Intelligent Engineering Systems}, vol.~20, no.~2, pp. 113--121, 2016. [Online]. Available: \url{https://journals.sagepub.com/doi/abs/10.3233/KES-160335}
\BIBentrySTDinterwordspacing

\bibitem{9363800}
T.~Liu, Y.~Yin, and X.~Yang, ``Research on logistics distribution routes optimization based on aco,'' in \emph{2020 5th International Conference on Information Science, Computer Technology and Transportation (ISCTT)}, 2020, pp. 641--644.

\bibitem{10075989}
R.~Wang, ``Research on cold chain logistics distribution path optimization based on improved ant colony algorithm,'' in \emph{2023 IEEE 3rd International Conference on Power, Electronics and Computer Applications (ICPECA)}, 2023, pp. 1250--1253.

\bibitem{9788179}
W.~Niu, ``A novel multiobjective optimization for tourism route based on improvement aco method and topology optimization,'' in \emph{2022 6th International Conference on Intelligent Computing and Control Systems (ICICCS)}, 2022, pp. 701--704.

\bibitem{10147562}
T.~Indravattana and T.~Silawan, ``Load balancing in multi depot pickup and delivery problem with ant colony optimization,'' in \emph{2023 8th International Conference on Business and Industrial Research (ICBIR)}, 2023, pp. 913--916.

\bibitem{10873146}
Z.~Jiang and Y.~Sun, ``Research on the planning of police patrol line based on ant colony optimization algorithm,'' in \emph{2024 International Seminar on Artificial Intelligence, Computer Technology and Control Engineering (ACTCE)}, 2024, pp. 74--76.

\bibitem{10462744}
D.~W. Sari, S.~Dwijayanti, and B.~Y. Suprapto, ``Path planning for an autonomous vehicle based on the ant colony algorithm: A review,'' in \emph{2023 International Workshop on Artificial Intelligence and Image Processing (IWAIIP)}, 2023, pp. 57--62.

\bibitem{surana2019benchmarking}
P.~Surana, ``Benchmarking optimization algorithms for capacitated vehicle routing problems,'' 2019.

\end{thebibliography}

\end{document}